\documentclass[letterpaper, 10 pt, conference]{ieeeconf}
\IEEEoverridecommandlockouts
\usepackage{cite}
\usepackage{amsmath,amssymb,amsfonts}
\usepackage[ruled,linesnumbered]{algorithm2e}
\DontPrintSemicolon
\usepackage{graphicx}
\usepackage{graphbox}
\usepackage{textcomp}
\usepackage{xcolor}
\usepackage{mathtools}
\usepackage{array}
\usepackage{float}
\usepackage{makecell}
\usepackage{kantlipsum} 
\usepackage{mwe} 
\usepackage{xurl} 
\usepackage{hyperref}
\usepackage{booktabs}
\usepackage{tikz}
\usepackage{multicol}
\usepackage{multirow}
\usepackage{adjustbox}
\usepackage{bm}
\usepackage{cases}
\usepackage{subcaption}

\DeclareMathAlphabet\mathbfcal{OMS}{cmsy}{b}{n}

\title{\LARGE \bf
Multi-Objective Trajectory Planning with Dual-Encoder
}

 \author{Beibei Zhang$^{1,*}$, Tian Xiang$^{1,*}$, Chentao Mao$^{2}$, Yuhua Zheng$^{1}$, \\Shuai Li$^{2}$, Haoyi Niu$^{1}$, Xiangming Xi$^{1}$, Wenyuan Bai$^{1}$, Feng Gao$^{1, \dag}$
 \thanks{$^{1}$ Beibei Zhang, Tian Xiang, Yuhua Zheng, Haoyi Niu, Xiangming Xi, Wenyuan Bai, and Feng Gao are with Zhejiang Lab, Hangzhou, China.}
 \thanks{$^{2}$ Chentao Mao and Shuai Li are with Hangzhou Institute for Advanced Study, University of Chinese Academy of Sciences, Hangzhou, China.}
 \thanks{$^*$ Beibei Zhang and Tian Xiang have equally contributed to this work.}
 \thanks{$\dag$ Feng Gao is the corresponding author.}
 }

\begin{document}
\maketitle
\begin{abstract}
Time-jerk optimal trajectory planning is crucial in advancing robotic arms' performance in dynamic tasks. Traditional methods rely on solving complex nonlinear programming problems, bringing significant delays in generating optimized trajectories. In this paper, we propose a two-stage approach to accelerate time-jerk optimal trajectory planning. Firstly, we introduce a dual-encoder based transformer model to establish a good preliminary trajectory. This trajectory is subsequently refined through sequential quadratic programming to improve its optimality and robustness. Our approach outperforms the state-of-the-art by up to 79.72\% in reducing trajectory planning time. Compared with existing methods, our method shrinks the optimality gap with the objective function value decreasing by up to 29.9\%. 
\end{abstract}

\section{Introduction}
Driven by the imperative for increased productivity in future smart factories, 
we demand robotic arms to efficiently and reliably accomplish 
a sequence of dynamic tasks. 
As an example, aircraft assembly lines employ robotic arms 
to place components with varying shapes and sizes, 
adjusting to the changing requirements of the assembly process~\cite{masehian2021assembly}. 
A robotic arm trajectory, commonly quoted as 
a robot configuration as a function of time, 
is pivotal for a robot manipulator to fulfill the tasks. 
To optimize productivity and enhance the stability of a robot manipulator, 
we consider time and the time derivative of acceleration, 
known as a jerk, as optimization objectives when proposing the trajectory.
Our time objective targets to minimize the overall duration of a trajectory 
and the jerk objective prioritizes diminishing the abrupt changes in trajectory accelerations. 

\begin{figure}[t]
	\centering
	\includegraphics[width=0.8\columnwidth]{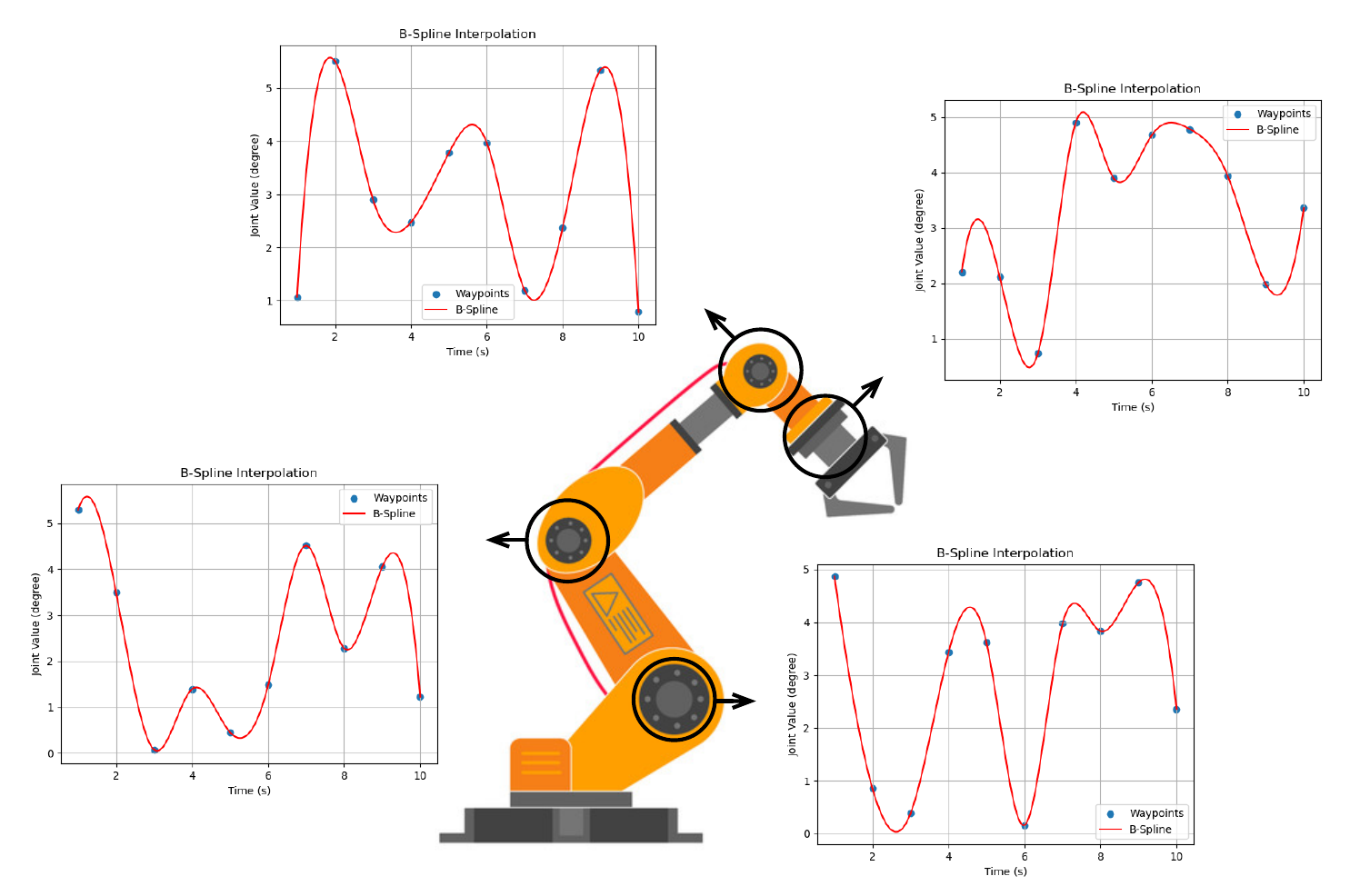}	
        \caption{We show a 4DOFs robot manipulator. A joint space trajectory composes waypoints for each joint and an interpolation function passes through the waypoints.}
	\label{fig:joint-trajectory}
\end{figure}

The trajectory can be carried out in the joint space 
with a series of successive joint values that 
each robot joint should assume over time, referred to as waypoints.
Trajectory planning seeks to create interpolation functions, 
such as polynomial~\cite{vsvejda2015interpolation}, 
Bezier~\cite{manyam2021trajectory}, and B-spline~\cite{huang2018optimal}, 
to accurately traverse the waypoints as illustrated in Fig.~\ref{fig:joint-trajectory}. 
Fundamentally, obtaining a time-jerk optimal trajectory 
amounts to solving a nonlinear programming problem, which is NP-hard~\cite{gasparetto2008technique}. 
The inherent computational demands of trajectory planning 
result in prolonged trajectory generation time, 
making it difficult to cope with the dynamic task requirements 
of future smart factories. 
In this paper, we focus on speeding up 
the time-jerk optimal trajectory planning process.

Current research addresses the time-jerk optimal trajectory 
planning through two classes of methods. 
\textbf{A first group of work} tackles the problem 
by optimizing a system of equations that consists of 
the interpolation function type and the constraints imposed on it. 
The problem can be solved through 
genetic algorithms (GA)~\cite{huang2018optimal, wang2022time, lin2018efficient} 
that iteratively improve the approximation set 
or the sequential quadratic programming (SQP) algorithm~\cite{zhang2021time} 
that decomposes the nonlinear and non-convex problem into a sequential convex problem. 
\textbf{Another line of research} resorts to devising supervised learning models, 
such as artificial neural networks~\cite{yin2019machine} and long short-term memory~\cite{molina2021trajectory}, 
trained on time-jerk optimal trajectory datasets. 
The supervised learning models take kinematic configurations 
of robotic manipulators and waypoints as inputs 
to generate a time-jerk optimal trajectory. 

Upon investigating and experimenting on 
the released code of state-of-the-art methods, 
we empirically observe that: 
\textbf{First}, the result optimality and computation time 
of the GA or the SQP methods are highly 
sensitive to how good the initial estimate is. 
As an example~\cite{gukov2022real}, 
the SQP-based solution takes an average 
2.6 seconds to generate a trajectory 
and requires multiple times of optimizations with different initial estimates 
to get a time-jerk optimal trajectory.
\textbf{Second}, although it is faster to generate a 
trajectory with supervised learning methods, 
the black-box nature of such approaches 
leads to poor robustness 
when faced with kinematic constraints 
and joint configurations not represented in the training data. 
This lack of robustness renders these methods unsuitable for future smart factories~\cite{masehian2021assembly}.

To sum up, the optimization-based methods 
fall short of delivering efficiency. 
This gap not only affects the throughput of production lines 
but also limits the potential for real-time adjustments, 
which are essential features in future smart factories. 
The supervised learning-based methods show promising efficiency that caters to the demands of future smart manufacturing. 
However, the methods struggle with robustness 
and fail to achieve optimal results.

In response to these challenges, 
our paper introduces a two-stage trajectory planning method. 
\textbf{First}, with the observation that 
a good initial estimate of 
the trajectory noticeably improves the convergence 
speed of the SQP optimization routine, 
we propose a dual-encoder based transformer model~\cite{vaswani2017attention} 
to predict initial trajectory parameters. 
\textbf{Next}, based on the initial parameters, we further optimize the trajectory with SQP to derive a time-jerk optimal trajectory, thereby enhancing the robustness of 
the method without compromising efficiency. 
Technically, we design a dual-encoder based  
transformer model that consists of a source encoder and a context encoder. 
The source encoder represents 
the dynamic characteristics of the current joint. 
Meanwhile, the context encoder depicts the 
motion relationships between the current joint and other joints.

Thereafter, we conduct extensive experiments on five counterparts 
including SQP~\cite{zhang2021time}, NSGA-II~\cite{huang2018optimal}, IPTP~\cite{pham2014general}, TOTG~\cite{kunz2013time}, and LSTM~\cite{molina2021trajectory}. 
Empirically, our approach consistently and significantly 
outperforms state-of-the-art methods. 
Interestingly, we observe that, 
unlike other methods, 
our approach is less impacted by trajectory length, 
measured by the number of waypoints, 
in terms of trajectory planning time.

To summarize, our key contributions are as follows:
1) A two-stage solution to expedite the generation of 
time-jerk optimal trajectory, where we design a dual-encoder based transformer model trained on time-jerk optimal 
trajectories to predict the initial estimate of the SQP.
2) Our method achieves new state-of-the-art performance over optimization-based approaches and supervised learning methods, particularly highlighting the balance between computational efficiency and robustness in trajectory planning. 
The implementations are released, hoping to facilitate future research in robust and efficient multi-objective trajectory planning.

\section{Problem Statement}

Suppose a robotic manipulator with $K$ degrees of freedom (DOFs) 
is employed to follow a path consisting of $I$ waypoints. 
The joint values of all joints are denoted as $\mathbfcal{Q} = \{\bm{q}^k\}_{k=1}^K$. 
Further, we describe the joint values of the $k$-th joint as $\bm{q}^k = \{q_1^k, \dots, q_i^k, \dots, q_I^k\}$, 
where $q_i^k$ represents the joint value of the $k$-th joint at the $i$-th waypoint. For each $k$-th joint, 
our objective is to propose a three-times differentiable function 
$\bm{q}^k(t)$ that depends on time $t \geq 0$ and interpolates $\{q_1^k, \dots, q_i^k, \dots, q_I^k\}$.
Building on recent research~\cite{artunedo2021jerk}, 
we represent $\bm{q}^k(t)$ using a 5th-order B-spline function $f(h^k,t)$, where $\bm{q}^k(t) = f(h^k,t)$. 
The set $h^k$ comprises coefficients and knot vectors associated with the B-spline of the $k$-th joint. 
Collectively, we denote the simultaneous joint values of $K$ joints 
at time $t$ as $\bm{q}(t) = \{\bm{q}^k(t)\}_{k=1}^K = f(\bm{h},t)$.

Suppose that the trajectory starts from time $0$ and $T$ is the timestamp of the last waypoint in the trajectory, 
where $\bm{q}^k(T) = q_I^k$ for all $k \in \{1, 2, \dots, K\}$. 
Given the joint limit $\bm{q}_{\text{max}}$, maximum velocity $\bm{\dot{q}}_{\text{max}}$, 
maximum acceleration $\bm{\ddot{q}}_{\text{max}}$ of the joints, 
and maximum jerk $\bm{\dddot{q}}_{\text{max}}$, 
$\bm{q}(t)$ is subject to satisfying kinematic constraints 
and boundary conditions listed in Eq.~\ref{eqn:constraints} and Eq.~\ref{eqn:boundary},


\begin{equation}\label{eqn:constraints}
\begin{cases}
   |\bm{q}(t)| < \bm{q}_{\text{max}}, &\\
   |\bm{\dot{q}}(t)| < \bm{\dot{q}}_{\text{max}}, &\\
   |\bm{\ddot{q}}(t)| < \bm{\ddot{q}}_{\text{max}}, &\\
   |\bm{\dddot{q}}(t)| < \bm{\dddot{q}}_{\text{max}},
\end{cases}
\forall t \in [0,T], 
\end{equation}

\begin{equation}\label{eqn:boundary}
\bm{\dot{q}}(0)=\bm{\dot{q}}(T)=\bm{\ddot{q}}(0)=\bm{\ddot{q}}(T)=0.
\end{equation}
The total jerk $\mathbfcal{J}$ of $K$ joints along the trajectory is defined as:
\begin{equation}
    \mathbfcal{J} = \sum_{k=1}^{K}\sqrt{\int_{0}^{T} 
    \bm{\dddot{q}}^k(t)^2 \,dt/T}
\end{equation}
With the above settings, we formulate the time-jerk optimal trajectory planning problem as
\begin{align}
\min_{\bm{h}} \ \ \ & \lambda J + (1-\lambda) T, \\
\text{s.t.} \ \ \   & \text{Eq.(1) and Eq. (2)},
\end{align}
where $\lambda$ is a balancing parameter.
The two objectives are inherently contradictory due to their opposing effects. 
Minimizing time tends to increase jerk fluctuation 
and disrupt the smoothness of the trajectory. 
Conversely, prioritizing jerk reduction can extend the execution time 
and compromise overall work efficiency. 

To expedite the optimal trajectory planning time, 
we propose a dual-encoder based transformer model trained on time-jerk optimal trajectories to predict the initial trajectory that will be further optimized by SQP. Next, we will delve into the details of the proposed model.

\begin{figure*}[t]
 \centering
 \begin{adjustbox}{width=0.85\textwidth}
    \includegraphics{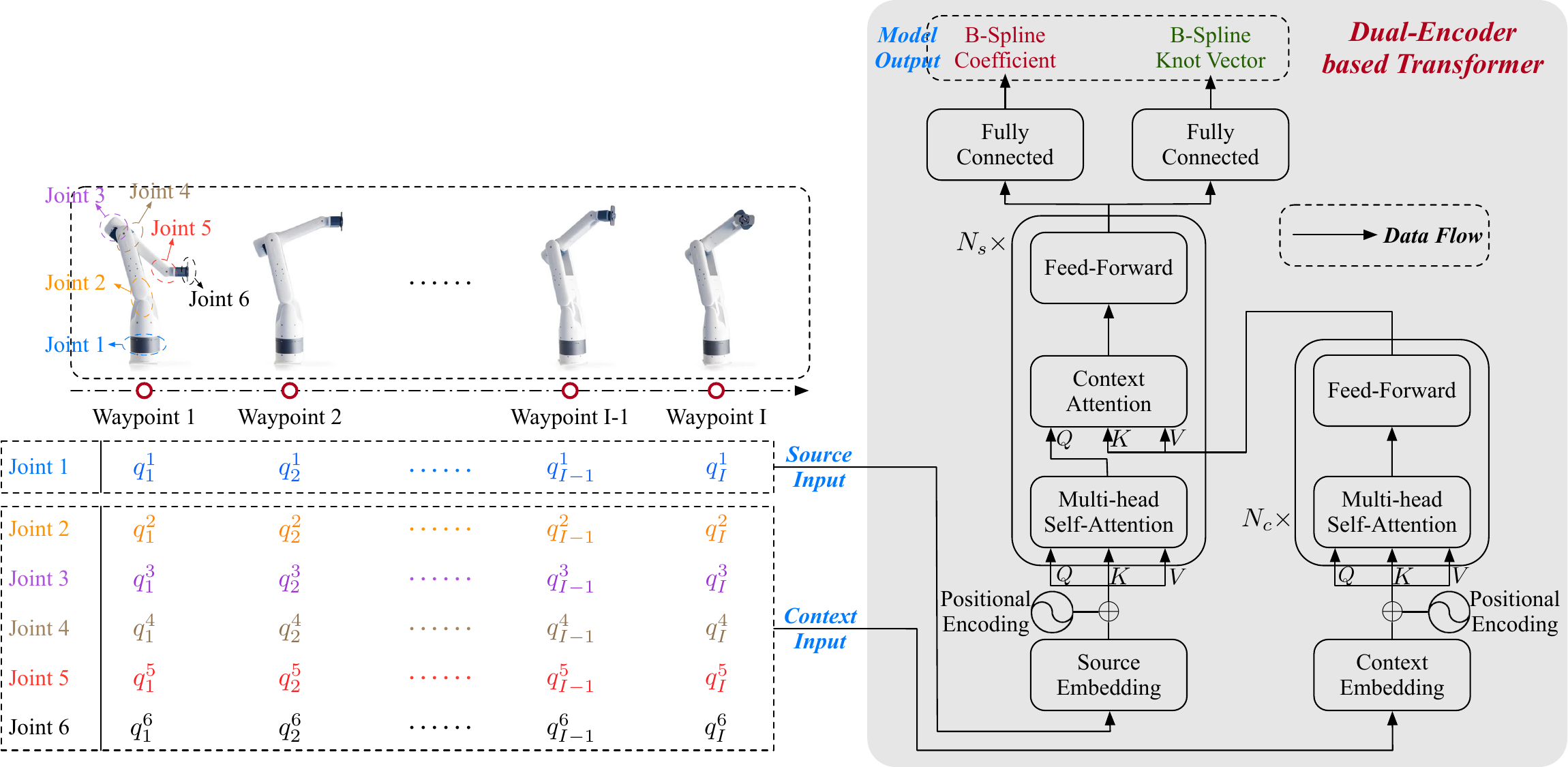}
  \end{adjustbox}
  \caption{Dual-Encoder based Transformer architecture.}
  \label{fig:transformer-model}
  \vspace{-1em}
\end{figure*}

\section{Proposed Method}
Fig.~\ref{fig:transformer-model} depicts the overall structure of the model that predicts the initial trajectory. 
Given the waypoints of $k$-th joint, we take the joint values $\mathbfcal{Q}_{src} = \bm{q}^k = \{q_1^k, \dots, q_i^k, \dots, q_I^k\}$ as the input to the source encoder. 
Typically, a robotic arm has multiple degrees of freedom, 
corresponding to multiple joints. 
The motion of each joint is constrained by other joints, 
confining its movement within a certain range. 
To capture the kinematic relations between different joints, 
we introduce a context encoder that extracts the kinematic information among joints. 
Specifically, we employ a multi-head self-attention encoder 
to compute the representation matrix of $\mathbfcal{Q}_{ctx}$, 
where $\mathbfcal{Q}_{ctx}=\{\bm{q}^1, \dots, \bm{q}^{k-1},\bm{q}^{k+1},\dots, \bm{q}^{K}\}$, 
consisting of the concatenation of the joint values of the other $K-1$ joints. 

Suppose the number of joint values in $\mathbfcal{Q}_{src}$ is $I$, 
then the number of joint values in $\mathbfcal{Q}_{ctx}$ is $(K-1)\times I$. 
To adapt to the structure of the Transformer, 
we propose an embedding layer with a dimension of $D=32$ 
to encode the joint values in $\mathbfcal{Q}_{src}$ and $\mathbfcal{Q}_{ctx}$. 
Consequently, we have $\mathbfcal{Q}_{src} \in \mathbb{R}^{I \times D}$, and $\mathbfcal{Q}_{ctx} \in \mathbb{R}^{(K-1)I \times D}$. The number of waypoints in different trajectories varies, causing a variable number of elements in $\mathbfcal{Q}_{src}$ and $\mathbfcal{Q}_{ctx}$. To address this, we obtain the maximum number of waypoints in our trajectory dataset, termed as $I_{max}$. Therefore, the number of elements in $\mathbfcal{Q}_{src}$ and $\mathbfcal{Q}_{ctx}$ is $L=(K-1)\times I_{max}$ that contains $I$, $(K-1)\times I$ joint values respectively. The number of padding elements in $\mathbfcal{Q}_{src}$ and $\mathbfcal{Q}_{ctx}$ are $(K-1)\times I_{max} - I$ and $(K-1)\times (I_{max} - I)$. Next, we incorporate positional encoding to characterize the relative orders of waypoints along the trajectory. We refer to the output of $\mathbfcal{Q}_{src}$ and $\mathbfcal{Q}_{ctx}$ after applying positional encoding as $\mathbfcal{Q}_{src\_emb}$ and $\mathbfcal{Q}_{ctx\_emb}$ respectively. We have $\mathbfcal{Q}_{src\_emb}\in \mathbb{R}^{L\times D}$ and $\mathbfcal{Q}_{ctx\_emb}\in \mathbb{R}^{L\times D}$.

\subsection{Context Encoder}
The context encoder comprises a stack of $N_c$ identical context encoding layers, 
each of which has two sublayers. The first sublayer operates as 
a multi-head self-attention mechanism (MultiHead), 
which allows the direct modeling of dependencies between every two joints. 
The operation of the first sublayer is given by 
\begin{equation}
\mathbfcal{Z}_c^{1}=\text{MultiHead}(\bm{Q}_c,\bm{K}_c,\bm{V}_c), 
\end{equation}
where $\bm{Q}_c=\bm{K}_c=\bm{V}_c=\mathbfcal{Q}_{ctx\_emb}$. 
That is 
\begin{equation}
    \mathbfcal{Z}_c^{1}=\text{MultiHead}(\mathbfcal{Q}_{ctx\_emb},\mathbfcal{Q}_{ctx\_emb},\mathbfcal{Q}_{ctx\_emb}).
\end{equation} 
The multi-head self-attention layer incorporates a residual connection 
and a layer normalization (LayerNorm) to enhance model training stability~\cite{vaswani2017attention}. 
We denote the output of the first sublayer as 
\begin{equation}
\mathbfcal{O}_c^{1}=\text{LayerNorm}(\mathbfcal{Z}_c^{1}+\mathbfcal{Q}_{ctx\_emb}). 
\end{equation}

We design the second sublayer as a position-wise feed-forward network (FFN), which takes the output of the multi-head attention sublayer as input and undergoes a nonlinear transformation through a two-layer fully connected network with a ReLU activation function. We describe the procedure as
\begin{equation}
\mathbfcal{F}_c^{1}=\text{FFN}(\mathbfcal{O}_{c}^{1})=\text{ReLU}(\mathbfcal{O}_{c}^{1}\mathbf{W}_{c1}^1+\mathbf{b}_{c1}^1)\mathbf{W}_{c2}^1+\mathbf{b}_{c2}^1, 
\end{equation}
where $\mathbfcal{O}_c^{1} \in \mathbb{R}^{L\times{D}}$ is the output of the first sublayer, $\mathbf{W}_{c1}^1, \mathbf{b}_{c1}^1 \text{ and } \mathbf{W}_{c2}^1, \mathbf{b}_{c2}^1$ represent the the first and the second layer parameters of the FFN. Likewise, we append a residual connection and a layer normalization to the second sublayer producing the output of the first context encoding layer, referred to as
\begin{equation}
\mathbfcal{C}^{1}=\text{LayerNorm}(\mathbfcal{F}_c^{1}+\mathbfcal{O}_c^{1}).    
\end{equation}

We repeat the above process $N_c$ times deriving
\begin{align}
\mathbfcal{Z}_c^{N_c} &= \text{MultiHead}(\mathbfcal{C}^{N_c-1},\mathbfcal{C}^{N_c-1},\mathbfcal{C}^{N_c-1}), \\
\mathbfcal{O}_c^{N_c} &= \text{LayerNorm}(\mathbfcal{Z}_c^{N_c}+\mathbfcal{C}^{N_c-1}), \\
\mathbfcal{F}_c^{N_c} &= \text{FFN}(\mathbfcal{O}_{c}^{N_c}) \nonumber \\ 
                      &=\text{ReLU}(\mathbfcal{O}_{c}^{N_c}\mathbf{W}_{c1}^{N_c}+\mathbf{b}_{c1}^{N_c})\mathbf{W}_{c2}^{N_c}+\mathbf{b}_{c2}^{N_c}, \\
\mathbfcal{C}^{N_c} &= \text{LayerNorm}(\mathbfcal{F}_c^{N_c}+\mathbfcal{O}_c^{N_c}), 
\end{align}
where $\mathbfcal{Z}_c^{N_c} \in \mathbb{R}^{L\times{D}}$ and $\mathbfcal{C}^{N_c} \in \mathbb{R}^{L\times{D}}$, 
which correspond to the hidden state and output of the $N_c$-th layer, respectively.

\subsection{Source Encoder and Context Integration}
Up to this point, we have presented the design of context encoder 
that extracts cues from $\mathbfcal{Q}_{ctx\_emb}$. 
Next, we introduce our source encoder 
with $N_s$ identical source encoding layers to encode $\mathbfcal{Q}_{src\_emb}$. 
Each source encoding layer comprises 
three consecutive sublayers: 
a multi-head self-attention layer, 
a context attention layer, 
and a feed-forward layer. 
Technically, we propose a multi-head self-attention layer 
to encapsulate the relative motion change among waypoints in a trajectory.
Formally, we apply the multi-head self-attention to the source embedding $\mathbfcal{Q}_{src\_emb}$: 
\begin{align}
\mathbfcal{Z}_s^{1} &=\text{MultiHead}(\mathbfcal{Q}_{src\_emb}, \mathbfcal{Q}_{src\_emb}, \mathbfcal{Q}_{src\_emb}), \\
\mathbfcal{O}_{s1}^{1} &=\text{LayerNorm}(\mathbfcal{Z}_s^{1}+\mathbfcal{Q}_{src\_emb}), 
\end{align}
where $\mathbfcal{O}_{s1}^{1}$ is the output of the multi-head self-attention layer.

To capture the kinematic relations between the current joint and other joints, 
we integrate the output of the context encoder 
into the source encoder with a context attention layer:
\begin{align}
\mathbfcal{G}_s^{1} &= \text{MultiHead}(\mathbfcal{O}_{s1}^{1},\mathbfcal{C}^{N_c},\mathbfcal{C}_c^{N_c}), \\
\mathbfcal{O}_{s2}^{1} &= \text{LayerNorm}(\mathbfcal{G}_s^{1}+\mathbfcal{O}_{s1}^{1}).
\end{align}

Then, the position-wise feed-forward network (FFN) is applied, which can be expressed as:
\begin{align}
\mathbfcal{F}_s^{1} &= \text{FFN}(\mathbfcal{O}_{s2}^{1}) \nonumber \\
                    &=\text{ReLU}(\mathbfcal{O}_{s2}^{1}\mathbf{W}_{s1}^1+\mathbf{b}_{s1}^1)\mathbf{W}_{s2}^1+\mathbf{b}_{s2}^1, \\
\mathbfcal{S}^{1} &= \text{LayerNorm}(\mathbfcal{F}_s^{1}+\mathbfcal{O}_{s2}^{1})
\end{align}

We iterate the same process $N_s$ times:
\begin{align}
\mathbfcal{Z}_s^{N_s} &= \text{MultiHead}(\mathbfcal{S}^{N_s-1},\mathbfcal{S}^{N_s-1},\mathbfcal{S}^{N_s-1}), \\
\mathbfcal{O}_{s1}^{N_s} &= \text{LayerNorm}(\mathbfcal{Z}_s^{N_s}+\mathbfcal{S}^{N_s-1}), \\
\mathbfcal{G}_s^{N_s} &= \text{MultiHead}(\mathbfcal{O}_{s1}^{N_s},\mathbfcal{C}^{N_c},\mathbfcal{C}_c^{N_c}), \\
\mathbfcal{O}_{s2}^{N_s} &= \text{LayerNorm}(\mathbfcal{G}_s^{N_s}+\mathbfcal{O}_{s1}^{N_s}), \\
\mathbfcal{F}_s^{N_s} &= \text{FFN}(\mathbfcal{O}_{s2}^{N_s}) \nonumber \\
    &= \text{ReLU}(\mathbfcal{O}_{s2}^{N_s}\mathbf{W}_{s1}^{N_s}+\mathbf{b}_{s1}^{N_s})\mathbf{W}_{s2}^{N_s}+\mathbf{b}_{s2}^{N_s}, \\
\mathbfcal{S}^{N_s} &=\text{LayerNorm}(\mathbfcal{F}_s^{N_s}+\mathbfcal{O}_{s2}^{N_s})
\end{align}
where $\mathbfcal{S}^{N_s} \in \mathbb{R}^{L\times{D}}$ is the representation of source joint $\mathbf{q}^{k}$ at the ${N_s}$-th layer.

To predict the coefficients and knot vector of 
a B-spline curve that fits a robotic arm trajectory, 
we link the source encoder's output 
to two distinct two-layer fully connected networks. 
This process yields separate outputs for each branch, 
enabling us to derive the predictions for the coefficients and knot vector as outlined below:
\begin{align}
\mathbfcal{R}_{coef}=\text{ReLU}(\mathbfcal{S}^{N_s}\mathbf{W}_{coef1}+\mathbf{b}_{coef1})\mathbf{W}_{coef2}+\mathbf{b}_{coef2}, \\
\mathbfcal{R}_{knot}=\text{ReLU}(\mathbfcal{S}^{N_s}\mathbf{W}_{knot1}+\mathbf{b}_{knot1})\mathbf{W}_{knot2}+\mathbf{b}_{knot2},   
\end{align}    
where $\mathbfcal{R}_{coef} \in \mathbb{R}^{1\times{M}}$, 
$\mathbfcal{R}_{knot} \in \mathbb{R}^{1\times{N}}$, $M$ denotes the number of coefficients and $N$ denotes the number of knot vectors.

\subsection{Training}
Our training data consists of the coefficients and knot vector of the B-splines that fit the time-jerk optimal trajectory. 
Our training objective is to minimize the error between the model output and the ground truth in the training dataset. 
Given that the output of the dual-encoder 
bifurcates into two branches, 
one predicting the coefficients of the B-spline 
and the other predicting the knot vector of the B-spline, 
we design two different loss functions to cater to two branches. 
The composite loss function can be described as:
\[
    L = \theta_1 \cdot L_{\text{coef}} + \theta_2 \cdot L_{\text{knot}}
\]
where $L_{\text{coef}}$ is the L1 smooth loss for the B-spline coefficients, $L_{\text{knot}}$ is the L1 loss for the knot vector, $\theta_1$ and $\theta_2$ are weighting parameters that balance the loss components.

For the branch that predicts the B-spline coefficients, 
we employ the L1 smooth loss function, which can be defined as:
\[
     L_{\text{coef}}(\mathbf{y}, \mathbf{\hat{y}}) = \frac{1}{N} \sum_{i=1}^{N} \text{smooth}_{L1}(y_i - \hat{y}_i)\
\]
where $\mathbf{y}$ represents the ground truth coefficient vector, $\mathbf{\hat{y}}$ represents the predicted coefficient vector, $N$ is the number of coefficients, and $\text{smooth}_{L1}$ is the smooth L1 loss function applied element-wise, referred to as:

\[
\text{smooth}_{L1}(x) = \begin{cases} 
0.5 \cdot x^2 & \text{if } |x| < 1 \\
|x| - 0.5 & \text{otherwise}
\end{cases}
\]

This choice is motivated by the need for a loss function that not only facilitates robustness but also smooths out predictions to avoid abrupt changes, which are critical for accurately modeling the trajectory's curvature. 

For the branch predicting the knot vector, we utilize the L1 loss function, which can be defined as:
\[L_{\text{knot}}(\boldsymbol{\eta}, \boldsymbol{\hat{\eta}}) = \frac{1}{M} \sum_{i=1}^{M} |\eta_i - \hat{\eta}_i|
\]
where $\boldsymbol{\eta}$ represents the ground truth knot vector, $\boldsymbol{\hat{\eta}}$ represents the predicted knot vector, and $M$ is the number of knots in the vector.

The adoption of the L1 loss for the knot vector prediction addresses the sparse changes in knot vectors, focusing on precise knot placement vital for trajectory fidelity.

We also employ a learning rate decay strategy to fine-tune model weights more precisely in later training stages.

  \begin{figure*}[t]
	\centering
	\begin{subfigure}[t]{0.48\textwidth}
            \centering
	      \includegraphics[width=\textwidth]{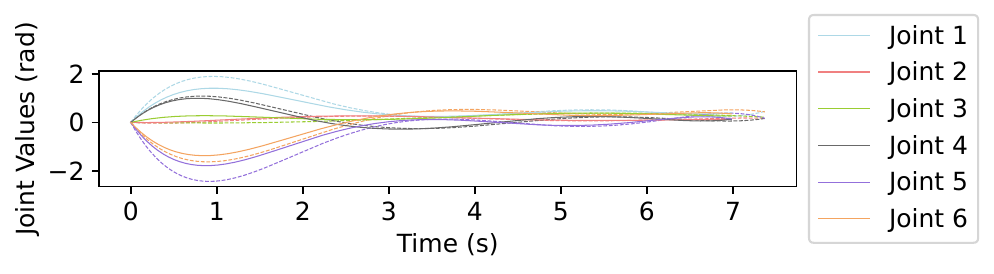}	
            \caption{Worst-case scenario.}
	   \label{fig:optimal-trajectory-worst}
	   \end{subfigure}
	\hfill
	\begin{subfigure}[t]{0.48\textwidth}
		\centering
		\includegraphics[width=\textwidth]{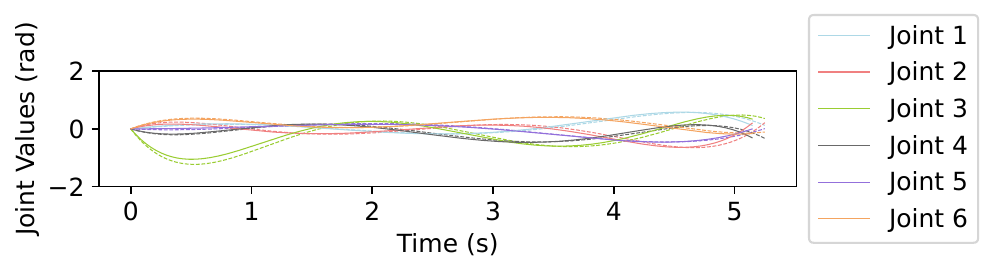}
		\caption{Best-case scenario.}
		\label{plot:optimal-trajectory-best}
	\end{subfigure}
	\caption{Comparison of trajectories generated by our dual-encoder model (dashed) to the ground truth generated by 5th-order BSpline (solid).}
	\label{fig:optimal-trajectory}
\end{figure*}

  \begin{figure*}[t]
	\centering
	\begin{subfigure}[t]{0.24\textwidth}
		\centering
		\includegraphics[width=\textwidth]{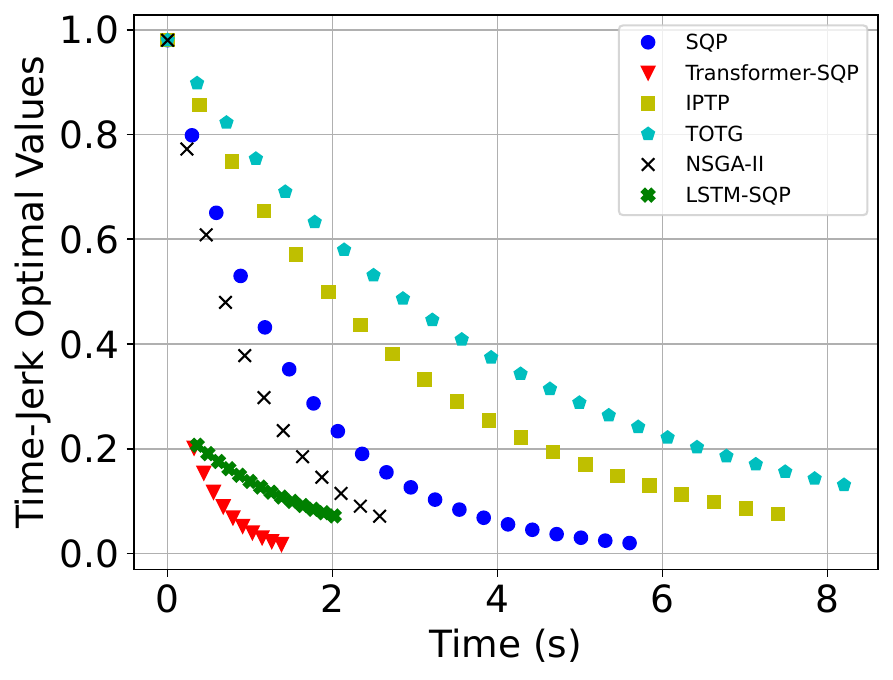}
		\caption{path length=6}
		\label{plot:pathlength6}
	\end{subfigure}
	\hfill
	\begin{subfigure}[t]{0.24\textwidth}
		\centering
		\includegraphics[width=\textwidth]{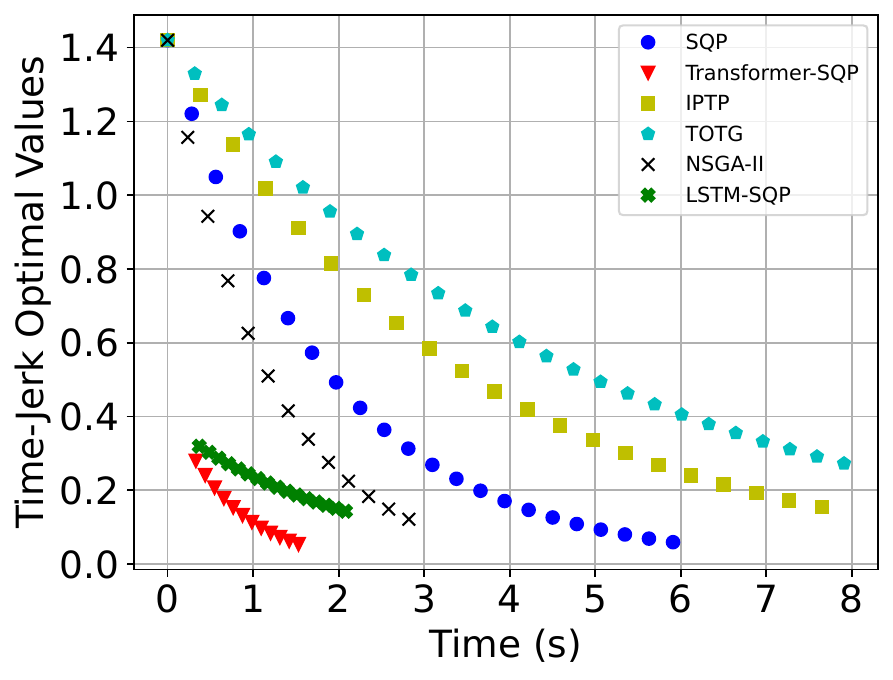}
		\caption{path length=12}
		\label{plot:pathlength12}
	\end{subfigure}
	\hfill
	\begin{subfigure}[t]{0.24\textwidth}
		\centering
		\includegraphics[width=\textwidth]{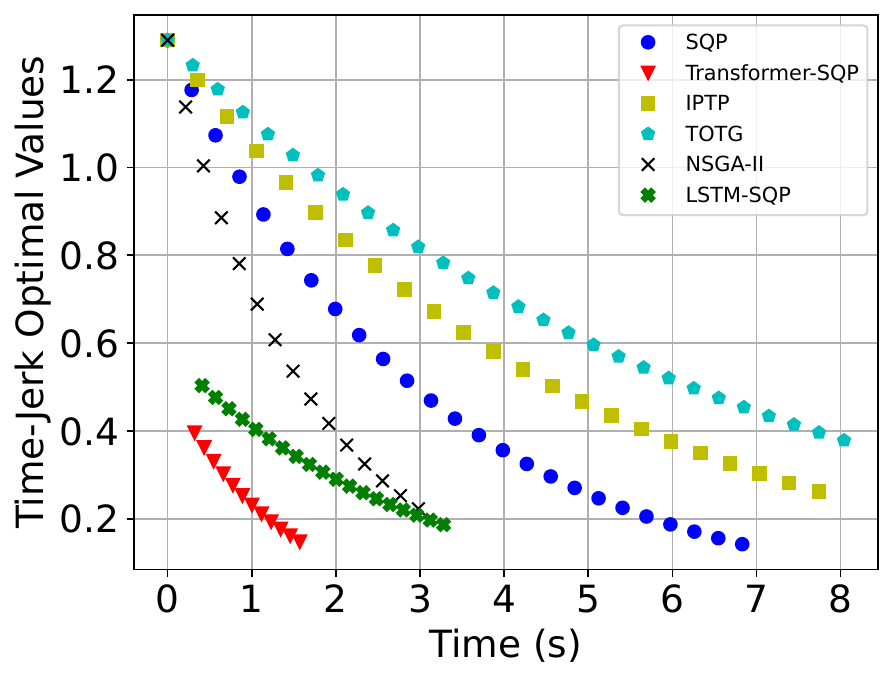}
		\caption{path length=24}
		\label{plot:pathlength24}
	\end{subfigure}
	\hfill
	\begin{subfigure}[t]{0.24\textwidth}
		\centering
		\includegraphics[width=\textwidth]{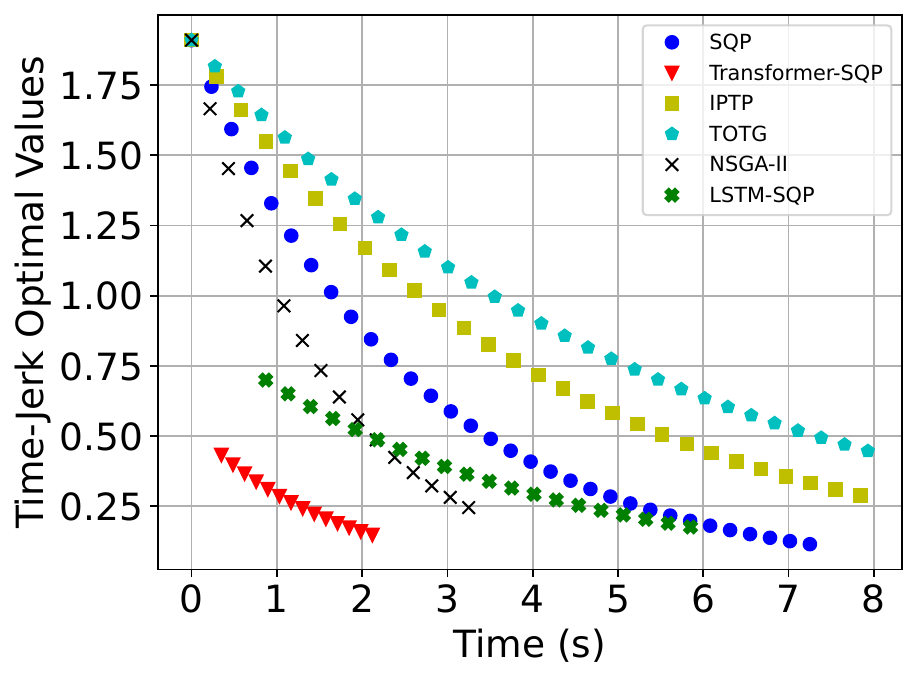}
		\caption{path length=48}
		\label{plot:pathlength48}
	\end{subfigure}
	\caption{Convergence trends of SQP, our method, IPTP, TOTG, NSGA-II, LSTM with various numbers of waypoints over time.}
	\label{plot:optimization-results}
\end{figure*}

\begin{figure}[t]
	\centering
	\begin{minipage}[t]{0.44\columnwidth}
		\centering
	    \includegraphics[width=\linewidth, align=t]{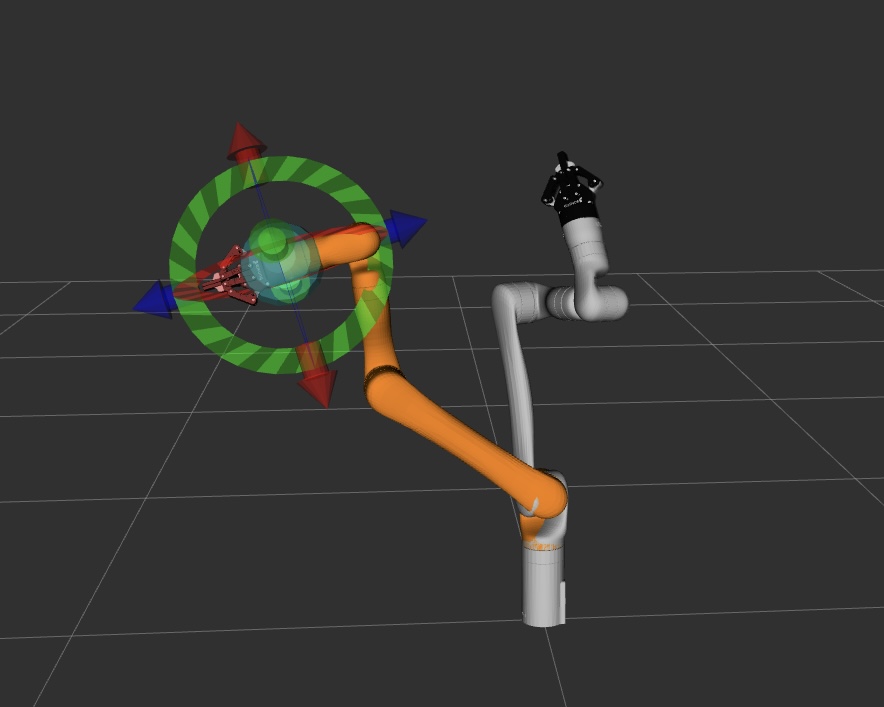}
            \vspace{0.75em}
	    \caption{Time-jerk optimal trajectory data collection with MoveIt2.}
	    \label{plot:moveit2}
	\end{minipage}
	\begin{minipage}[t]{0.54\columnwidth}
		\centering
	    \includegraphics[width=\linewidth, align=t]{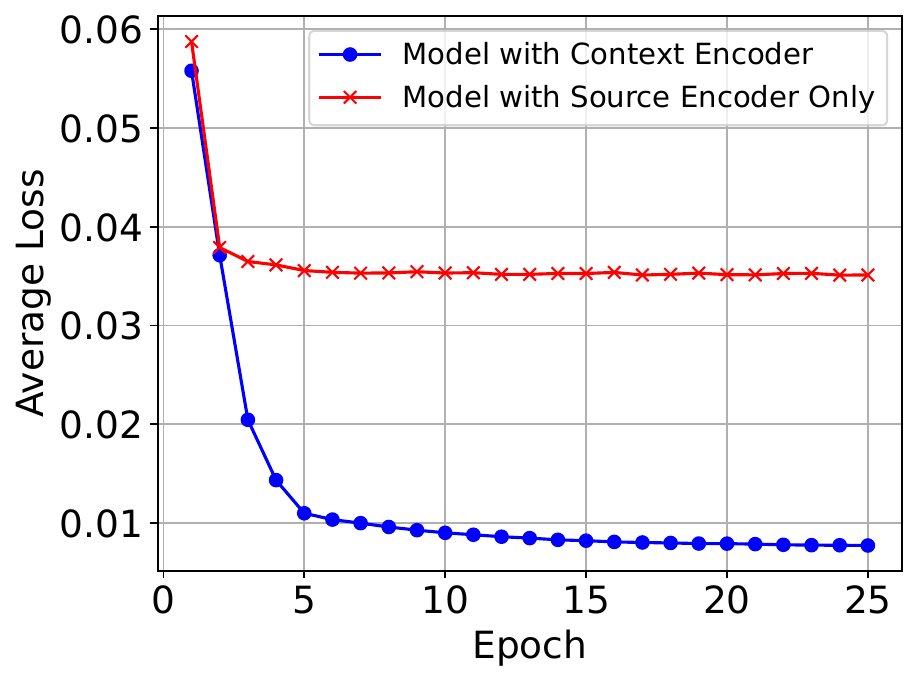}
            \vspace{-0.5em}
	    \caption{Comparison of training loss between our method and source encoder-only model.}
	    \label{fig:loss-trend}
	\end{minipage}
\end{figure}

  \begin{figure}[t]
	\centering
	\begin{subfigure}[t]{0.49\columnwidth}
		\centering
		\includegraphics[width=\textwidth]{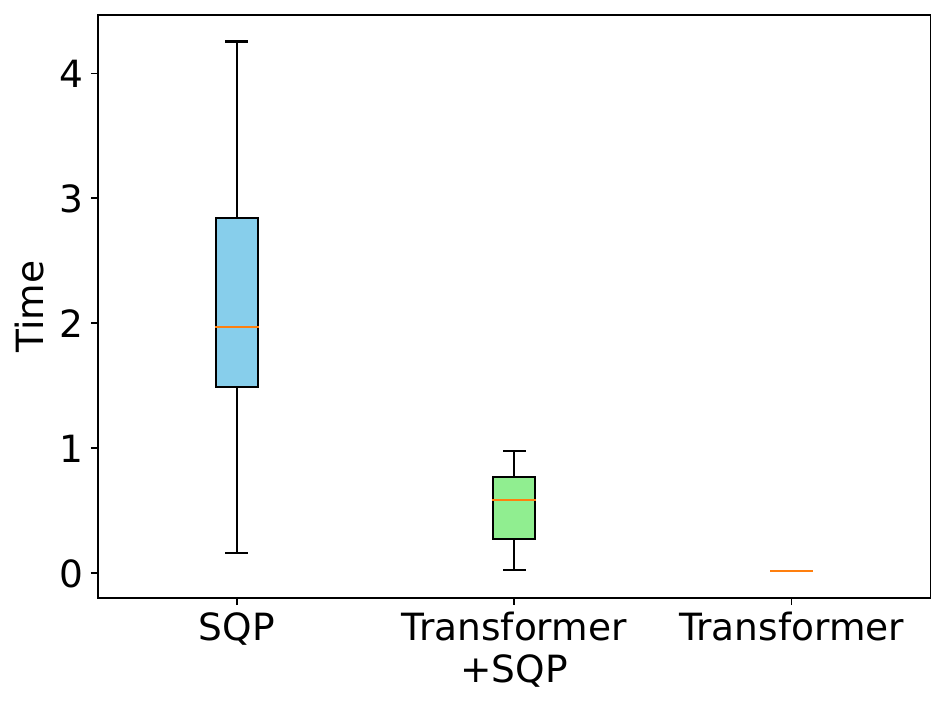}
		\caption{path length=6}
		\label{plot:box6}
	\end{subfigure}
	\begin{subfigure}[t]{0.49\columnwidth}
		\centering
		\includegraphics[width=\textwidth]{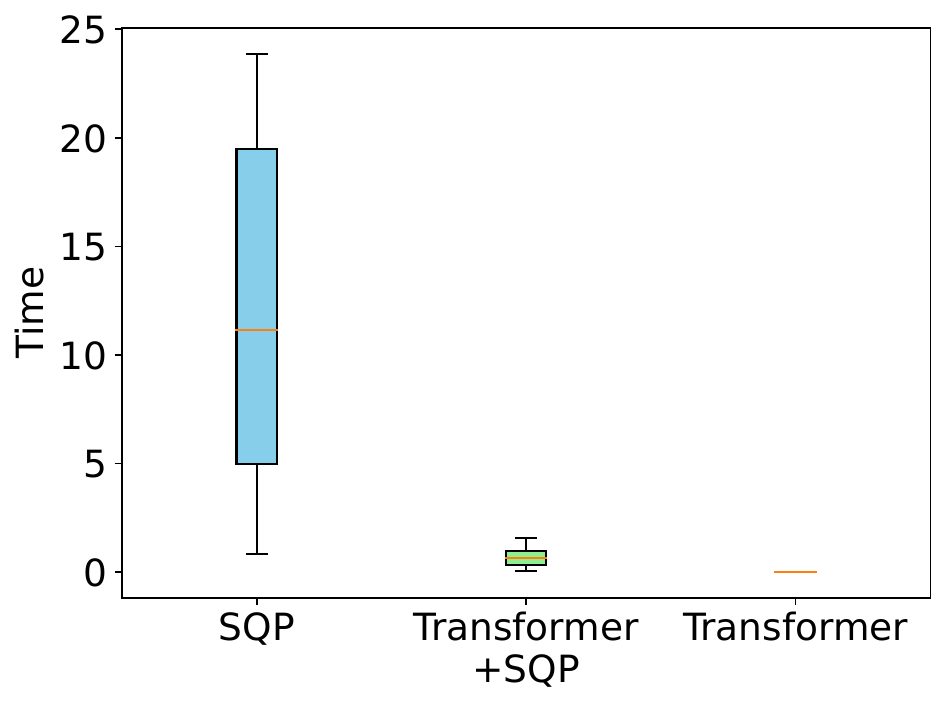}
		\caption{path length=12}
		\label{plot:box12}
	\end{subfigure}
	\begin{subfigure}[t]{0.49\columnwidth}
		\centering
		\includegraphics[width=\textwidth]{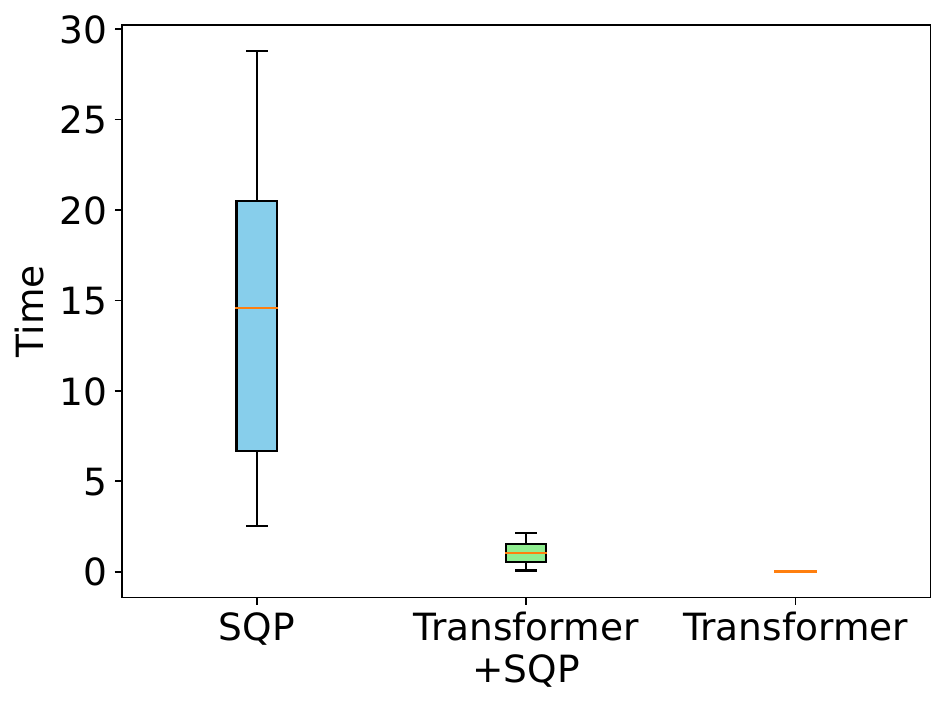}
		\caption{path length=24}
		\label{plot:box24}
	\end{subfigure}
	\begin{subfigure}[t]{0.49\columnwidth}
		\centering
		\includegraphics[width=\textwidth]{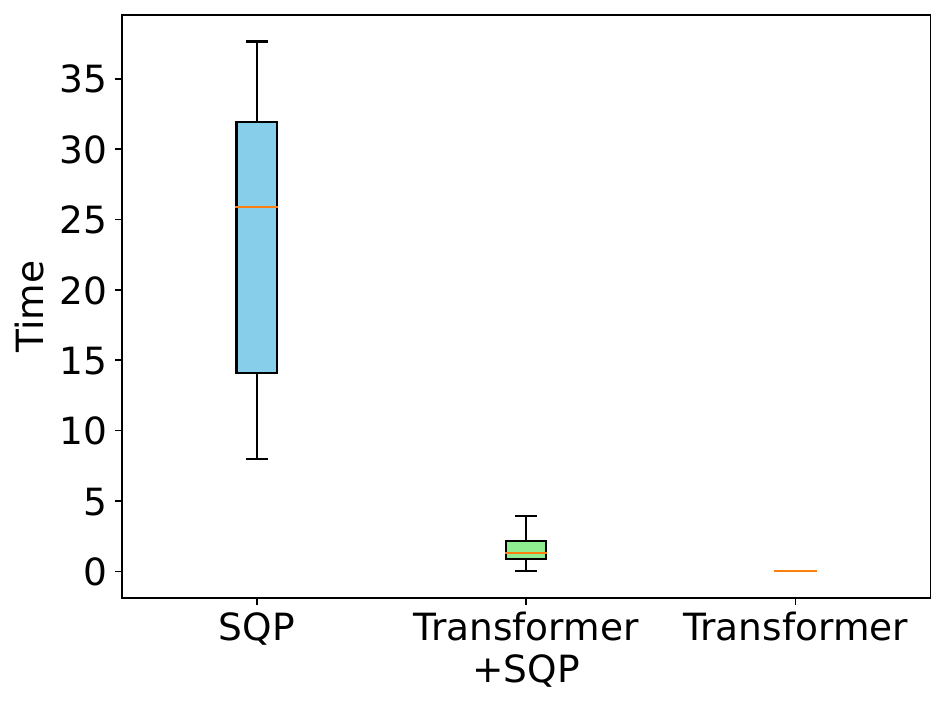}
		\caption{path length=48}
		\label{plot:box48}
	\end{subfigure}
	\caption{Time variations (seconds) comparison among SQP, Transformer+SQP, and Transformer across different path lengths: a boxplot analysis.}
	\label{plot:boxplot results}
    \vspace{-1em}
\end{figure}

 \section{Experiment}
 \subsection{Setup}
\textbf{Datasets.} We leverage a Kinova Gen3 6DOFs 
 robotic arm to generate time-jerk optimal trajectories 
 in Moveit2 as depicted in Fig.~\ref{plot:moveit2}. 
 The dataset consists of trajectories with lengths 
 ranging from 6 to 48 waypoints. 
 These waypoints are uniformly distributed in the 
 Cartesian space that falls within the manipulator's 
 operational range, with inverse kinematics 
 employed to derive the corresponding joint values. 
 We collect a total of 80532 trajectories, 
 where 56372 trajectories are used for training, 
 16106 trajectories are used for validation, 
 and 7784 trajectories are used for testing.
 Every trajectory data contains six data items, 
 each of which contains the joint values of a Kinova Gen3 joint at the waypoints 
 and the joint values of the other Kinova Gen3 joints at the waypoints. 
 We feed the joint values of the current joint to the source encoder 
 and the joint values of the other joints to the context encoder.
 We train and serve the dual-encoder based transformer model on 
 an NVIDIA TITAN RTX GPU. 

 \textbf{Training Configs.} The dual-encoder model features a 32-dimensional 
 embedding for joint values, 8 attention heads, 
 and 6 stacked layers to enhance learning capabilities. 
 It incorporates a dropout rate of 0.1 to prevent over-fitting. 
 We employ an Adam optimizer~\cite{kingma2014adam} with the learning rate set to 0.001 and incorporate a weight decay of 1e-4 for regularization.
 An adaptive learning rate scheduler, $\mathtt{ReduceLROnPlateau}$, 
 is used to adjust the learning rate based on performance.

 We showcase the performance of our proposed model 
 with a worst-case scenario and a best-case scenario from our test data.
 As illustrated in Fig.~\ref{fig:optimal-trajectory}, 
 we plot the ground truth trajectory B-splines and the B-splines generated 
 by our model.
 In the worst-case scenario, 
 the generated B-spline exhibits a joint value error margin of less 
 than 23.47\% in comparison to the ground truth.
 In the best-case scenario, 
 the generated B-spline exhibits a joint value error margin of less 
 than 5.26\% in comparison to the ground truth.

 \textbf{Counterparts.} We benchmark our approach against several established algorithms, including SQP~\cite{zhang2021time}, NSGA-II~\cite{huang2018optimal}, Iterative Parabolic Time Parametrization (IPTP)~\cite{pham2014general}, Time-Optimal Trajectory Generation (TOTG)~\cite{kunz2013time}, and LSTM~\cite{molina2021trajectory}. NSGA-II starts with a preliminary feasible set, iteratively enhancing it through a process that involves sorting solutions based on non-domination, selection, crossover, and mutation operations to generate new solutions while ensuring compliance with velocity and acceleration constraints. IPTP adjusts time intervals based on the most constraining velocity across joints, incrementally modifying them until acceleration constraints are met. TOTG smoothens paths with circular blends, establishes a maximum velocity profile in the phase plane, and accomplishes time parameterization through numerical integration. LSTM is utilized for trajectory planning by first generating paths with the Rapidly Exploring Random Tree (RRT) method, then refining these paths using LSTM networks trained on environmental data. We take the output of LSTM as an initial value for SQP optimization in our experiment. The next subsection highlights the efficiency of our proposed method in trajectory planning and its superiority in reducing the objective function values.
 
 \subsection{Results}
In our experiments, we conduct an ablation study by removing the context encoder, leaving only the source encoder in the transformer model. As depicted in Fig.~\ref{fig:loss-trend}, our proposed method has achieved a loss reduction to 0.01 after just 5 epochs of training and further converged to below 0.007 within 15 epochs. In contrast, the loss of the source encoder only model plateaued at 0.035 after 5 epochs, demonstrating inferior performance in comparison to the dual-encoder model. 
This demonstrates the effectiveness of the context encoder. 

 Fig.~\ref{plot:optimization-results} illustrates the comparison of time consumed by various optimization algorithms in solving the time-jerk optimal trajectory planning problem, with path lengths of 6, 12, 24, and 48. All methods commence optimization from identical initial values. However, both Transformer+SQP and LSTM+SQP incorporate a preliminary inference step to refine the initial values before further optimization through SQP. In contrast, the other methods proceed the optimization process directly on the original initial values. This analysis selects a set of results that balance time efficiency and trajectory optimality. 
 It can be observed that, after obtaining the initial estimate through our well-trained transformer model and then carrying out subsequent optimizations, the optimal trajectory planning time improves up to 77.01\%, 79.72\%, 80.47\%, 47.32\%, and 52.13\% in comparison to SQP, IPTP, TOTG, NSGA-II, and LSTM+SQP respectively. In terms of results optimality, the proposed methods outperform SQP, IPTP, TOTG, NSGA-II, and LSTM+SQP by up to 8.43\%, 16.71\%, 22.2\%, 29.9\%, and 29.7\% respectively. 
 From the above analysis, we empirically observed that LSTM+SQP performance is greatly affected by the path length, whereas our proposed model keeps its performance across varying path lengths.

 Furthermore, we employ a boxplot analysis to compare the time variations across different path lengths for SQP, Transformer+SQP, and Transformer, as shown in Fig.~\ref{plot:boxplot results}. Due to SQP's high dependency on initial values, it exhibits the greatest fluctuation in processing time, with a difference of up to 29.64 seconds between the maximum and minimum values at the path length of 48. In contrast, the Transformer+SQP approach, benefiting from accurate initial values inferred by our dual-encoder based Transformer, demonstrates a more stable time variation as path length increases. Notably, the Transformer model shows the shortest inference time, appearing almost as a constant line in the figure. This indicates the efficiency of our proposed method to predict initial values for SQP optimization and highlights its ability to maintain performance consistency across different trajectory lengths.
 
\section{Conclusion}
In this article, we have proposed a two-stage approach to address the challenges associated with time-jerk optimal trajectory planning for robotic arms. Initially, a dual-encoder based transformer model, trained on datasets of time-jerk optimized trajectories, is utilized to quickly generate initial values. These initial values serve as a foundation for further refinement through sequential quadratic programming optimization, leading to enhanced trajectory planning. Our extensive experimental evaluation validates the effectiveness of this method, demonstrating its potential to improve the efficiency and performance of multi-objective trajectory planning. 

\bibliographystyle{IEEEtran}
\bibliography{main}

\end{document}